\documentclass{article}
\usepackage{spconf,amsmath,graphicx}
\usepackage{amssymb}
\usepackage{spconf,graphicx}
\usepackage{algorithm}
\usepackage[noend]{algorithmic} 
\usepackage{verbatim}
\usepackage{amsmath}
\usepackage{amssymb}
\usepackage{url}
\usepackage{booktabs}
\usepackage{todonotes}

\newcommand\blfootnote[1]{
  \begingroup
  \renewcommand\thefootnote{}\footnote{#1}%
  \addtocounter{footnote}{-1}%
  \endgroup
}

\newtheorem{definition}{Definition}[section]
\newtheorem{theorem}{Theorem}[section]
\newtheorem{proof}{Proof}[section]

\DeclareMathOperator{\pop}{pop}
\DeclareMathOperator{\add}{add}
\DeclareMathOperator{\isEmpty}{isEmpty}

\title{Streaming Binary Sketching based on Subspace Tracking and Diagonal Uniformization}
    
\name{Anne Morvan$^{\star \dagger}$ \qquad Antoine Souloumiac$^{\star}$ \qquad C\'edric Gouy-Pailler$^{\star}$ \qquad Jamal Atif$^{\dagger}$}

\address{$^{\star}$ CEA, LIST, 91191 Gif-sur-Yvette, France \\
    $^{\dagger}$Universit\'e Paris-Dauphine, PSL Research University, CNRS, LAMSADE, 75016 Paris, France
   }

\begin{document}
%
\maketitle
\begin{abstract}
In this paper, we address the problem of learning compact similarity-preserving embeddings for massive high-dimensional \emph{streams} of data in order to perform efficient similarity search. We present a new online method for computing binary compressed representations -\emph{sketches}- of high-dimensional real feature vectors. Given an expected code length $c$ and high-dimensional input data points, our algorithm provides a $c$-bits binary code for preserving the distance between the points from the original high-dimensional space. Our algorithm does not require neither the storage of the whole dataset nor a chunk, thus it is fully adaptable to the streaming setting. It also provides low time complexity and convergence guarantees. 
We demonstrate the quality of our binary sketches through experiments on real data for the nearest neighbors search task in the online setting.
\end{abstract}
\begin{keywords}
Sketching, streaming, subspace tracking, Givens rotations\blfootnote{A. Morvan is partly supported by the DGA (French Ministry of Defense).}\blfootnote{Copyright 2018 IEEE. Published in the IEEE 2018 International Conference on Acoustics, Speech, and Signal Processing (ICASSP 2018), scheduled for 15-20 April 2018 in Calgary, Alberta, Canada. Personal use of this material is permitted. However, permission to reprint/republish this material for advertising or promotional purposes or for creating new collective works for resale or redistribution to servers or lists, or to reuse any copyrighted component of this work in other works, must be obtained from the IEEE. Contact: Manager, Copyrights and Permissions / IEEE Service Center / 445 Hoes Lane / P.O. Box 1331 / Piscataway, NJ 08855-1331, USA. Telephone: + Intl. 908-562-3966.}
\end{keywords}
%
\section{Introduction}
\label{sec:intro}

In large-scale machine learning applications such as computer vision or metagenomics, learning similarity-preserving binary codes is critical to perform efficient indexing of large-scale high-dimensional data. Storage requirements can be reduced and similarity search sped up by embedding high-dimensional data into a compact binary code.
A classical method is Locality-Sensitive Hashing (LSH)~\cite{Andoni2008NHA} for nearest neighbors search: 1) input data with high dimension $d$ are projected onto a lower $c$-dimensional space through a $c \times d$ random projection with i.i.d. Gaussian entries, 2) then a hashing function is applied to the resulting projected vectors to obtain the final binary codes.
Two examples for the hashing function are cross-polytope LSH~\cite{terasawa2007} which returns the closest vector from the set $\{\pm 1 e_{i}\}_{1 \leq i \leq c}$ where $\{e_{i}\}_{1 \leq i \leq c}$ stands for the canonical basis, and hyperplane LSH~\cite{indyk2015}, the $\operatorname{sign}$ function applied pointwise. To reduce the storage cost of the projection matrix and the  matrix-vector products computation times ($O(c \times d$)), a structured pseudo-random matrix can be used instead~\cite{indyk2015, bojarski17a} with a reduced time complexity of $O(d \log c)$ thanks to fast Hadamard and Fourier transforms.

In the context of nearest neighbors search or classification, the accuracy of the data sketches can be improved by learning this projection from data~\cite{SH08, SHL12, ITQ2013, IsoHash2012, CBEICML14}. As Principal Component Analysis (PCA) is a common tool for reducing data dimensionality, data are often projected onto the first $c$ principal components. But PCA alone is not sufficient. Indeed, the $c$ first principal components are chosen with a decreasing order of explained variance: principal directions with higher variance carry more information. Thus, associating each of the $c$ directions to one of the $c$ bits is equivalent to giving more weights to less informative directions and will lead to poor performance of the obtained sketches.     
To remedy this problem, after data have been projected on the first principal components of the covariance matrix, a solution consists in applying a suitable rotation on the projected data before performing the hashing function: it balances variance over the principal components. In work from~\cite{Jegou2010}, a random rotation is used giving quite good results. In ITerative Quantization (ITQ)~\cite{ITQ2013} or in Isotropic Hashing (IsoHash)~\cite{IsoHash2012}, the rotation is rather learned. In ITQ, the rotation is iteratively computed by solving an orthogonal Procustes problem: the alternating minimization of the quantization error of mapping data to the vertices of the $2^c$ hypercube. This technique is currently the state-of-the-art for computing similarity-preserving binary codes based on PCA.
ITQ and IsoHash come with a major drawback, though. They are completely offline since the whole dataset needs to be stored for computing the $c$ principal components. This can be prohibitive when dealing with lots of high-dimensional data.

\textbf{Contributions}: We introduce a streaming algorithm with convergence guarantees where data is seen only once and principal subspace plus the balancing rotation are updated as new data is seen. To obtain the principal subspace and the rotation, this requires additionally only the storage of two $c\times c$ matrices, instead of the whole initial and projected datasets as for ITQ or IsoHash. Our algorithm outperforms the known state-of-the-art online unsupervised method Online Sketching Hashing (OSH)~\cite{LengWC0L15} while being far less computationally demanding. 

\section{Related work}
Two paradigms exist to build hash functions~\cite{surveyHashing16}: data-independent~\cite{Andoni2008NHA, SKLSH09, KLSH11} and data-dependent methods. The latter ones learn the hash codes from a training set and perform better. The learning can be unsupervised~\cite{SH08,Liu11hashingwith,ITQ2013,IsoHash2012, Lee2012SphericalH,DisGraphHashingNIPS2014,CBEICML14,Raziperchikolaei16} aiming at preserving distances in the original space or (semi-)supervised which also tries to preserve label similarity~\cite{SHL12, supervisedHK12}. Some recent hashing functions involve deep learning~\cite{LaiPLY15, Chen2015CNN}. When the dataset is too large to be loaded into memory, distributed~\cite{HDDleng15} and online hashing techniques~\cite{Huang2013OH,LengWC0L15,CakirHBS17} have been developed. Online Hashing (OKH)~\cite{Huang2013OH} learns the hash functions from a stream of similarity-labeled pair of data with a "Passive-Aggressive" method. 
Recent approach from~\cite{CakirHBS17} relies on Mutual Information. Although claimed to be unsupervised, it is supervised as similarity labels are also needed between pairs of data.
In Online Sketching Hashing (OSH)~\cite{LengWC0L15}, the binary embeddings are learned from a maintained sketch of the dataset with a smaller size but preserving the property of interest. The proposed algorithm belongs to this latter category of online unsupervised hyperplanes-based hashing methods.

\section{The proposed online unsupervised model for binary quantization}

\subsection{Notations and problem statement}

We have a stream of $n$ data points $\{ x_t \in \mathbb{R}^d\}_{1 \leq t \leq n}$ supposed to be zero-centered. 
The goal is to have $b_t = \operatorname{sign}(\tilde{W}_t x_t) \in \{-1, 1\}^{c}$ for $t = 1 \ldots n$ where $c$ denotes the code length, $c \ll d$, s.t. for each bit $k = 1 \ldots c$, the binary encoding function is defined by $h_k(x_t) = \operatorname{sign}(\tilde{w}^T_{k,t} x_t)$ where $\tilde{w}_{k,t}$ are column vectors of hyperplane coefficients and $\operatorname{sign}(x) = 1$ if $x \geq 0$ and $-1$ otherwise which is applied component-wise on coefficients of vectors. So $\tilde{w}_{k,t}^T$ is a row of $\tilde{W}_t$ for each $k$, with $\tilde{W}_t \in \mathbb{R}^{c \times d}$. 
$b_t$ is computed and returned before $x_{t+1}$ is seen by using $x_t$ and $\tilde{W}_{t-1}$ solely. We consider in this paper the family of hash functions $\tilde{W}_t$ s.t. $\tilde{W}_t = R_t W_t$ where $W_t$ is the linear dimension reduction embedding applied to data and $R_t$ is a suitable $c \times c$ orthogonal matrix. 
Here, we take $W_t$ as the matrix whose row vectors $w_{k,t}^T$ are the $c$ first principal components of the covariance matrix $\Sigma_{X,t} = X_t X_t^T$ where $X_t$ denotes the dataset seen until data $t$. So the challenge is in tracking the principal subspace in an online fashion as new data is seen and in defining an appropriate orthogonal matrix\footnote{In the sequel, we use equally the term orthogonal matrix or rotation.} to rotate the projected data onto this principal subspace.
 
Works from~\cite{ITQ2013,Jegou2010} argued that defining an orthogonal transformation applied to the PCA-projected data which tends to balance the variances over the PCA-directions improves the quality of the hashing but $R$ was first learned to uniformize the variances for different directions only in work from~\cite{IsoHash2012} with Isotropic Hashing (IsoHash). In the latter, the problem is explicitly described as learning a rotation which produces isotropic variances of the PCA-projected dimensions. We propose here a new simpler online method UnifDiag for learning a rotation to uniformize the diagonal of the covariance of projected data after tracking the principal subspace $W_t$ from the data stream with Fast Orthonormal PAST (Projection Approximation and Subspace Tracking)~\cite{OPAST2000}, also named OPAST. At each iteration $t$, OPAST guarantees the orthonormality of $W_t$ rows and costs only $4dc + O(c^2)$ flops while storing only $W_t$ and a $c \times c$ matrix.
The rotation learning is completely independent from the principal subspace tracking.
Indeed, any other method for online PCA~\cite{FengXY13,YangX15a} or subspace tracking~\cite{OPAST2000} can be plugged before UnifDiag.

\subsection{UnifDiag: the proposed diagonal uniformization-based method for learning a suitable rotation}

Let the $c \times c$ symmetric matrix $\Sigma_{V,t} = V_t V_t^T$ be the covariance matrix of projected data seen until data $t$ $V_t = W_tX_t$.
$R_t$ is learned for each $t$ to balance the variance over the $c$ directions given by the $c$ principal components of $\Sigma_{V,t}$. $\Sigma_{V,t}$ is easy to update dynamically and we adapt OPAST algorithm to perform this while computing $W_t$. In the sequel, for clarity we drop the subscript $t$.
Let us consider the $c$ diagonal coefficients of $\Sigma_{V}$: $\sigma^2_1, ..., \sigma^2_c$ s.t. $\sigma^2_1 \geq ... \geq \sigma^2_c$. As $\Sigma_V$ is symmetric, $\operatorname{Tr}(\Sigma_V) = \sum_{i = 1}^c \sigma^2_i = \sum_{i = 1}^c \lambda_i$ where $\operatorname{Tr}$ stands for the Trace application and $\lambda_1, ..., \lambda_c$ are the $c$ first eigenvalues of $\Sigma_X$ s.t. $\lambda_1 \geq ... \geq \lambda_c$\footnote{If $W$ is exactly the $c$ first eigenvectors of $\Sigma_X$ -- for instance, if $W$ is obtained through PCA --, then $\forall i \in \{1, ..., c\}$, $\sigma_i^2 = \lambda_i$.}. Balancing variance over the $c$ directions can be seen as equalizing the diagonal coefficients of $\Sigma_V$ s.t. $\sigma^2_1 =  ... = \sigma^2_c \overset{def}{=} \tau$.
Since $\Sigma_V$ is symmetric, $\operatorname{Tr}(R \ \Sigma_V R^T) = \operatorname{Tr}(\Sigma_V)$. So in order to have $R \ \Sigma_V R^T$ with equal diagonal coefficients, we should set $\tau = \operatorname{Tr}(\Sigma_V) / c$.  
So, similarly to IsoHash, we formulate the problem of finding $R$ as the problem of equalizing the diagonal coefficients of $\Sigma_V$ to the value $\tau = \operatorname{Tr}(\Sigma_V) / c$.   
Our proposed optimal orthogonal matrix $R$ is built as a product of $c-1$ Givens rotations $G(i, j, \theta)$ described by Def.~\ref{def:givens}.

\begin{definition} \label{def:givens}
A Givens rotation $G(i,j, \theta)$ is a matrix of the form: 
\begin{center}
$G(i,j, \theta) = \begin{bmatrix}
1 & \cdots & 0 & \cdots & 0 & \cdots & 0 \\
\vdots & \ddots &\vdots & & \vdots & & \vdots \\ 
0 & \cdots & c & \cdots & -s & \cdots & 0 \\
\vdots & & \vdots & \ddots & \vdots & & \vdots \\ 
0 & \cdots & s & \cdots & c & \cdots & 0 \\
\vdots & & \vdots & & \vdots & \ddots & \vdots \\ 
0 & \cdots & 0 & \cdots & 0 & \cdots &1
\end{bmatrix}$ 
\end{center}
where $c = \cos(\theta)$ and $s = \sin(\theta)$ are at the intersections of the $i$-th and $j$-th rows and columns. The nonzero elements are consequently: $\forall k \neq i, \, j, \ g_{k,k} =1$, $g_{i,i} = g_{j,j} = c$, $g_{j,i} = -s$ and $g_{i,j} = s$ for $i > j$.
\end{definition}

The computation of $R$ follows the iterative Jacobi eigenvalue algorithm known as diagonalization process~\cite{Golub2000}:
\begin{align}
\Sigma_V & \leftarrow G(i,j, \theta) \ \Sigma_V \ G(i,j, \theta)^T \label{eq:updateCov} \\
R &\leftarrow R \ G(i,j, \theta)^T. 
\end{align}
Note that left (resp. right) multiplication by $G(i, j, \theta)$ only mixes $i$-th and $j$-th rows (resp. columns). The update from Eq.~\ref{eq:updateCov} only modifies $i$-th and $j$-th rows and columns of $\Sigma_V$. The two updated diagonal coefficients $(i,i)$ and $(j,j)$ only depend on $\Sigma_{V,i,i}$, $\Sigma_{V,j,j}$, $\Sigma_{V,j,i}$ and $\theta$ which reduces the optimization of $\theta$ to a $2$-dimensional problem, a classical trick when using Givens rotations~\cite{Golub2000}. Then we have Th.~\ref{thm:givens} by defining: $a \overset{def}{=} \Sigma_{V,j,j}$, $d \overset{def}{=} \Sigma_{V,i,i}$, $b \overset{def}{=} \Sigma_{V,j,i} = \Sigma_{V,i,j}$,
\begin{equation} \label{eq:2dimproblem}
\begin{pmatrix}
a' & b' \\
b' & d' 
\end{pmatrix}
\overset{def}{=}
\begin{pmatrix}
c & -s \\
s & c 
\end{pmatrix}
\begin{pmatrix}
a & b \\
b & d 
\end{pmatrix}
\begin{pmatrix}
c & s \\
-s & c 
\end{pmatrix}.
\end{equation}

\begin{theorem} \label{thm:givens}
If $\operatorname{min}(a,d) \leq \tau \leq \operatorname{max}(a,d)$ (sufficient condition)\footnote{Th.~\ref{thm:givens} uses only a sufficient condition. A weaker necessary and sufficient one to guarantee $|c_2| \leq 1$ and $s_2 \in [0,1]$ is $ \frac{a+d}{2} - \sqrt{ \left(\frac{a-d}{2}\right)^2 + b^2} \leq \tau \leq \frac{a+d}{2} + \sqrt{ \left(\frac{a-d}{2}\right)^2 + b^2}$.} then there exists one $\theta \in [ -\pi/2, \pi/2]$ s.t. $a' = \tau$, $d' = a + d - \tau$ and $b' = -s_2 \sqrt{ \left(\frac{a-d}{2} \right)^2 + b^2}$ with 
$\cos(\theta) = \sqrt{\frac{1 + c_1c_2 - s_1s_2}{2}}$ and $\sin(\theta) = - \frac{c_1 s_2 +c_2 s_1}{2 \cos \theta}$,
$c_1  = \left( \frac{a-d}{2} \right) / \sqrt{ \left(\frac{a-d}{2}\right)^2 + b^2}$, 
$s_1 = b / \sqrt{ \left(\frac{a-d}{2}\right)^2 + b^2}$,
$c_2 = (\tau - \frac{a+d}{2}) / \sqrt{ \left(\frac{a-d}{2}\right)^2 + b^2}$ and
$s_2  = \sqrt{1 - c_2^2} \in [0,1]$.
\end{theorem}
\begin{proof}
With $u(\theta) \overset{def}{=} \begin{pmatrix} \cos(2 \theta) &-\sin(2\theta)
\end{pmatrix}^T$, Eq.~\ref{eq:2dimproblem} gives:
$a' = \frac{a+d}{2} + 
\begin{pmatrix}
\frac{a-d}{2} & b
\end{pmatrix} . \ u(\theta)^T$, \\
$d' = \frac{a+d}{2} - \begin{pmatrix}
\frac{a-d}{2} & b
\end{pmatrix} . \ u(\theta)^T$ and
$b' =  \begin{pmatrix}
\frac{a-d}{2} & b
\end{pmatrix} . \begin{pmatrix}
\sin(2 \theta) \\
\cos(2\theta)
\end{pmatrix}$.
As the Givens angle $\theta$ should be parameterized s.t. all diagonal coefficients are set to $\tau$,
$\begin{pmatrix}
\frac{a-d}{2} & b
\end{pmatrix} . \ u(\theta)^T = \tau - \frac{a+d}{2}$. 
Previously defined $c_1$, $c_2$, $s_1$, $s_2$ and the condition $\operatorname{min}(a,d) < \tau < \operatorname{max}(a,d)$ gives $|c_2| \leq 1$ and $s_2 \in [0,1]$. Then the last equation becomes:
$\begin{pmatrix}
c_1 & s_1
\end{pmatrix} . \ u(\theta)^T = c_2$. 
A clear solution is:
$\ u(\theta)^T = \begin{pmatrix}
c_1c_2 -s_1s_2 \\
c_1s_2 + c_2 s_1 
\end{pmatrix}.$
Thus, one can take:
$\cos(\theta) = \sqrt{ \frac{1 + \cos(2 \theta)}{2}} = \sqrt{\frac{1 + c_1c_2 - s_1s_2}{2}} $,
$\sin(\theta) = \frac{\sin(2\theta)}{2 \cos(\theta)} = - \frac{c_1 s_2 +c_2 s_1}{2 \cos \theta}$ 
and the associated Givens rotation gives:
$a' = \tau$, \\
$d' = a + d - \tau$, \,
$b' = \begin{pmatrix}
\frac{a-d}{2} & b
\end{pmatrix} \begin{pmatrix}
\sin(2 \theta) \\
\cos(2\theta)
\end{pmatrix}$ \\ 
$= \sqrt{ \left(\frac{a-d}{2} \right)^2 + b^2} \begin{pmatrix}
c_1 & s_1 \end{pmatrix} \begin{pmatrix}
-c_1s_2 -c_2s_1 \\
c_1c_2  -s_1 s_2 
\end{pmatrix}$ \\
$= -s_2 \sqrt{ \left(\frac{a-d}{2} \right)^2 + b^2}$.
\end{proof}
Note that there is no need to compute explicitly $\theta$, $\theta_1$ or $\theta_2$. We now briefly describe the underlying diagonal uniformization algorithm. 
The mean of $\Sigma_V$ diagonal coefficients being equal to $\tau$, these indices sets are not empty:
$iInf \overset{def}{=} \{ \ l \in \{1, \ ..., \ c \} \ | \ \Sigma_{V,l,l} < \tau \}$ and $iSup \overset{def}{=} \{ \ l \in \{1, \ ..., \ c \} \ | \ \Sigma_{V,l,l} > \tau \}$.
Taking one index $j$ from $iInf$ and the other one $i$ from $iSup$ guarantees the condition of Th.~\ref{thm:givens}, which allows to set $\Sigma_{V,j,j}$ to the value $\tau$. The index $j$ can then be removed from $iInf$ and as $\Sigma_{V,i,i}$ is set to $a+d - \tau$, the index $i$ reassigned to $iInf$ if $\tau > \frac{a+d}{2}$ or in $iSup$ if $\tau < \frac{a+d}{2}$. The number of diagonal coefficients of $\Sigma_V$ different from $\tau$ has been decreased by one. Finally, the necessary number of iterations  to completely empty $iInf$ and $iSup$, i.e. uniformizing $\Sigma_V$ diagonal, is bounded by $c-1$. The method is summarized in Algorithm~\ref{algo:unif_diag} where $\pop(list)$ and $\add(list, e)$ are subroutines to delete and return the first element of $list$, resp. to add $e$ in $list$. 

\begin{algorithm}
\caption{Diagonal Uniformization algorithm (UnifDiag)} \label{algo:unif_diag}
\begin{algorithmic}[1]
\STATE \textit{Inputs} : $\Sigma_V$ ($c \times c$, symmetric), tolerance: $tol$ \\ 
\STATE $R \small{\gets} I_c$ // $c \times c$ Identity matrix; $\tau \small{\gets} \operatorname{Tr}(\Sigma_V) / c$  ; \,
$it = 0$
\STATE $iInf = \{ \ l \in \{1, \ ..., \ c \} \ | \ \Sigma_{V,l,l} < \tau- tol \}$  
\STATE $iSup = \{ \ l \in \{1, \ ..., \ c \} \ | \ \Sigma_{V,l,l} > \tau + tol \}$
\WHILE{ $it < c-1$ $\&$  not $\isEmpty(iInf)$ $\&$ not $\isEmpty(iSup)$}
\STATE \textit{// Givens rotation parameters computation}: 
\STATE $j \small{\gets} pop(iInf)$; 
$i \small{\gets}\pop(iSup)$; 
$a \small{\gets} \Sigma_V[j,j] $; 
$b \small{\gets} \Sigma_V[i,j]$; 
$d \small{\gets} \Sigma_V[i,i] $; $c$, $s$ (Th.~\ref{thm:givens});
$it \small{\gets} it + 1$\\
\STATE \textit{// $\Sigma_V$ update}: 
\STATE $row_j \small{\gets} \Sigma_V[j, :]$; \, $row_i \small{\gets} \Sigma_V[i, :]$
\STATE $\Sigma_V[j, :] = c \times row_j - s \times row_i$; 
\STATE $\Sigma_V[i, :] = s \times row_j + c \times row_i$  
\STATE $\Sigma_V[:, j] = \Sigma_V[j, :]$; \, 
$\Sigma_V[:, i] = \Sigma_V[i, :]$
\STATE $\Sigma_V[j,j] = a'$; 
$\Sigma_V[i,i] = d'$ ; 
$\Sigma_V[j,i] = b'$ (Th.~\ref{thm:givens})
\STATE \textit{// Rotation update}: 
\STATE $col_j \small{\gets} R[:, j]$; \,
$col_i \small{\gets} R[:, i]$
\STATE $R[:, j] = c \times col_j - s \times col_i$
\STATE $R[:, i] = s \times col_j + c \times col_i$ \\
\STATE \textit{ // Indices list update}: 
\IF{ $ \frac{a+d}{2} < \tau - tol$}
\STATE $\add(iInf, i)$
\ENDIF
\IF{ $ \frac{a+d}{2} > \tau + tol$}
\STATE $\add(iSup, i)$
\ENDIF
\ENDWHILE
\RETURN R
\end{algorithmic}
\end{algorithm}

\section{Complexity analysis of existing works}

Our algorithm requires the storage of two $c \times c$ matrices besides $W_t$ and $R_t$ obviously: one with OPAST to obtain $W_t$ and $\Sigma_{V,t}$ for $R_t$. 
One update with OPAST for $W_t$ and $\Sigma_{V,t}$ costs $4dc + O(c^2)$.
Then, to compute $R_t$, at most $c-1$ Givens rotations are needed, each implying four column or row multiplications i.e. $4c$ flops. So the final time complexity of our algorithm is $4dc + O(c^2)$.

We compare here the spatial and time costs of our method with, to the best of our knowledge, the only online unsupervised method: \textbf{Online Sketching Hashing} (OSH)~\cite{LengWC0L15} which is the most similar to ours, i.e. unsupervised, hyperplanes-based and reading one data point at a time. Despite what is announced, OSH is fundamentally mini-batch: the stream is divided into chunks of data for which a matrix $S \in \mathbb{R}^{d \times l}$ as a sketch of the whole dataset $X \in \mathbb{R}^{d \times n}$ is maintained. Then the principal components are computed from the updated sketch $S$. The projection of data followed by the random rotation can be applied only after this step. Therefore there are actually two passes over the data by reading twice data of each chunk.  
Without counting the projection matrix and the rotation, OSH needs spatially to maintain the sketch $S$ which costs $O(d \times l)$ with $c \ll l \ll d$. The SVD decomposition then needs $O(dl + l^2)$ space. In comparison, we only need $O(c^2)$.
For each round, OSH takes $O(dl^2 + l^3)$ time to learn the principal components, i.e. $O(dl + l^2)$ for each new data seen. 
We also compare with \textbf{IsoHash}~\cite{IsoHash2012}. Although it counts as an offline method because no technique is proposed to approximatively estimate the principal subspace, IsoHash rotation can be applied after for instance OPAST. 
IsoHash rotation computation involves an integration of a differential equation using Adams-Bashforth-Moulton PECE solver which costs $O(c^3)$ time.  
Even if $c$ is small in comparison to $d$ and the complexities do not either depend on $n$, our model has the advantages to have a lower time cost and to be much more simple than IsoHash. 
Thus, our method shows advantages in terms of spatial and time complexities over OSH and IsoHash. Moreover, binary hash codes can be directly computed as new data is seen, while OSH, as a mini-batch method, has a delay.

\section{Experiments}
\begin{figure}[htb]
\centering
\begin{minipage}[b]{0.4\linewidth}
  \centering
    \centerline{\includegraphics[width=5.8cm]{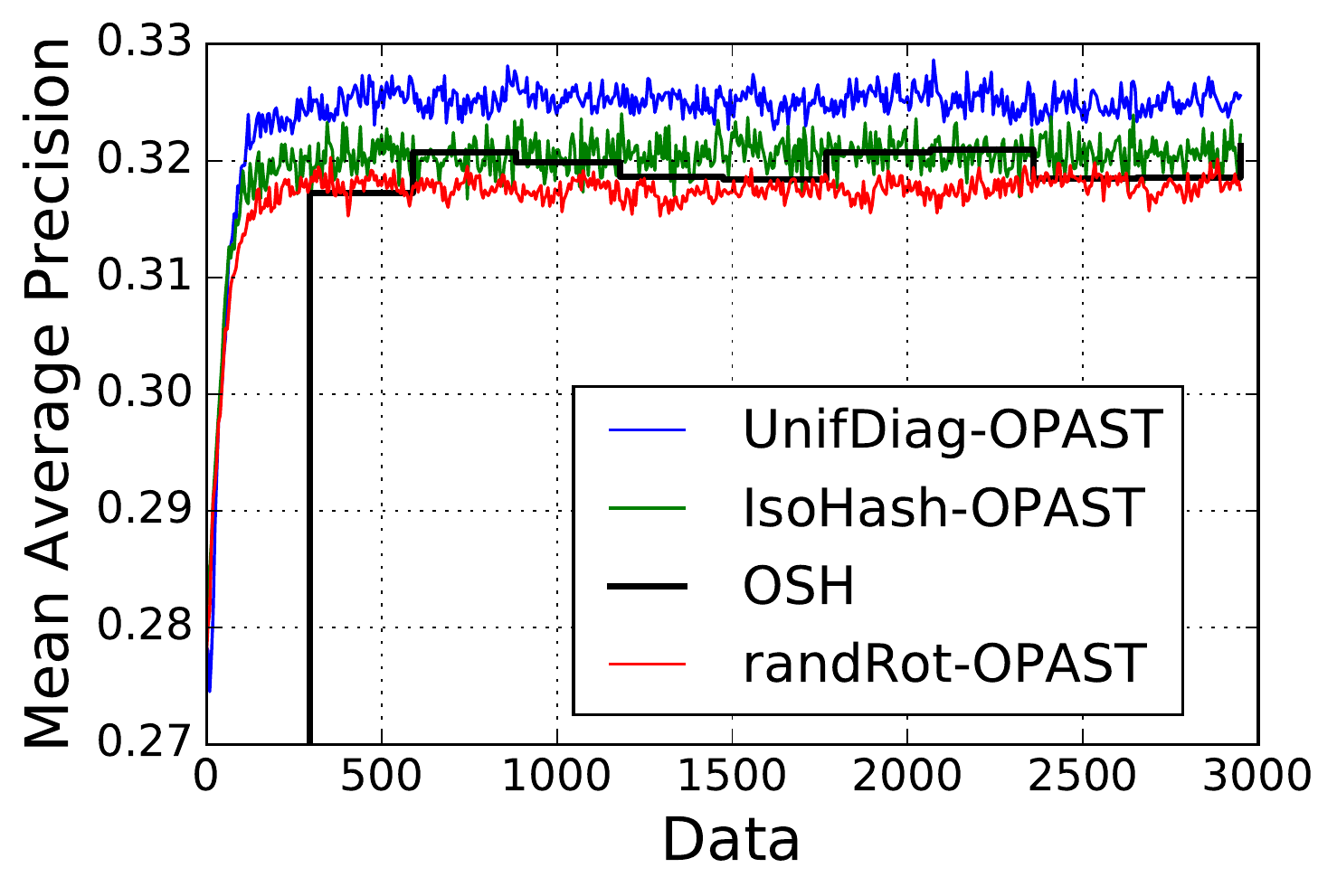}}
\end{minipage}
\vspace*{-0.21in}
\caption{mAP for $c = 32$ on CIFAR-10.}
\label{fig:CIFAR-10map_onlineStepFunction}
\end{figure}

\begin{figure}[htb]
\centering
\begin{minipage}[b]{0.4\linewidth}
  \centering
  \centerline{\includegraphics[width=5.8cm]{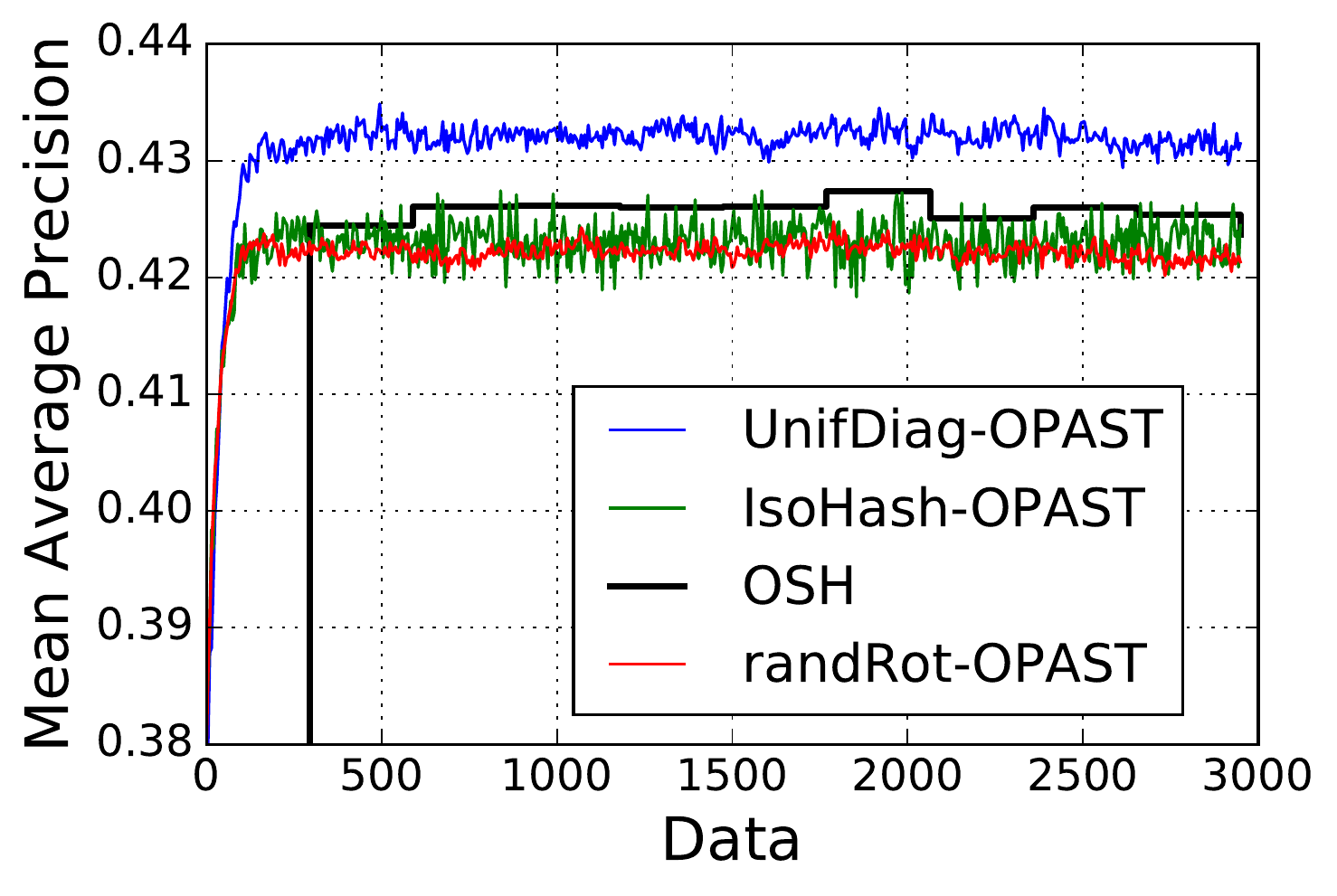}}
\end{minipage}
\vspace*{-0.21in}
\caption{mAP for $c = 32$ on GIST-1M.}
\label{fig:GIST1M_map_onlineStepFunction}
\end{figure}

Experiments are made on CIFAR-10 (\url{http://www.cs.toronto.edu/~kriz/cifar.html}) and GIST1M sets (\url{http://corpus-texmex.irisa.fr/}).
CIFAR-10 contains $60000$ $32 \times 32$ color images equally divided into 10 classes. 960-D GIST descriptors were extracted from data. GIST1M contains $1$ million $960$-D GIST descriptors, from which $60000$ instances were randomly chosen from the first half of the learning set. 
Quality of hashing has been assessed on the nearest neighbor (NN) search task performed on the binary codes instead of the initial descriptors. 
A nominal threshold of the average distance to the $50$th nearest neighbor is computed and determines the sets of neighbors and non-neighbors called Euclidean ground truth. $1000$ queries were randomly sampled and the remaining data are used as training set. 
We compared our method to three online baseline methods that follow the basic hashing scheme $\Phi(x_t) = \operatorname{sgn}(\tilde{W}_t x_t)$, where the projection matrix $\tilde{W}_t \in \mathbb{R}^{c \times d}$ is determined according to the chosen method:
1) \textbf{OSH} 
2) \textbf{RandRot-OPAST}: $W_t$ is the PCA matrix obtained with OPAST and $R_t$ a constant random rotation.
3) \textbf{IsoHash-OPAST}: $R_t$ is obtained with IsoHash. 
4) \textbf{UnifDiag-OPAST}: $R_t$ is obtained with UnifDiag. 
For OSH, the number of chunks is set to $200$ and $l = 50$.
Fig.~\ref{fig:CIFAR-10map_onlineStepFunction} and \ref{fig:GIST1M_map_onlineStepFunction} (best viewed in color) show the Mean Average Precision (mAP)~\cite{IR08} for both datasets for $c = 32$ (similar results are obtained for $c \in [8, 16, 64]$) averaged over $5$ random training/test partitions. 
Our algorithm outperforms all the compared methods. Code on GitHub: \url{annemorvan/UnifDiagStreamBinSketching/}.

\section{Conclusion}
We introduced a novel method for learning distance-preserving binary embeddings of high-dimensional data streams with convergence guarantees. Unlike classical state-of-the-art methods, our algorithm does not need to store the whole dataset and enables to obtain without delay a binary code as a new data point is seen. 
Our approach shows promising results as evidenced by the experiments. It can achieve better accuracy than state-of-the-art online unsupervised methods while saving considerable computation time and spatial requirements. 
Besides, the Givens rotations, that are a classical tool for QR factorization, singular and eigendecomposition or joint diagonalization, can also be used for uniformizing the diagonal of a symmetric matrix via an original Givens angle tuning technique.
Further work would be to investigate whether another rotation, not uniformizing the diagonal of the covariance matrix of the projected data, could be more optimal. Another interesting perspective is to evaluate the performance of the compact binary codes in other machine learning applications: instead of using the original data, one could use directly these binary embeddings to perform unsupervised or supervised learning while preserving the accuracy.

\vfill\pagebreak

\bibliographystyle{IEEEbib}
\bibliography{refs}

\begin{thebibliography}{10}

\bibitem{Andoni2008NHA}
A.~Andoni and P.~Indyk,
\newblock ``Near-optimal hashing algorithms for approximate nearest neighbor in
  high dimensions,''
\newblock {\em Commun. ACM}, , no. 1, pp. 117--122, 2008.

\bibitem{terasawa2007}
K.~Terasawa and Y.~Tanaka,
\newblock ``Spherical {LSH} for approximate nearest neighbor search on unit
  hypersphere,''
\newblock in {\em WADS}, 2007, pp. 27--38.

\bibitem{indyk2015}
A.~Andoni, P.~Indyk, I.~Laarhoven, T.and~Razenshteyn, and L.~Schmidt,
\newblock ``Practical and optimal {LSH} for angular distance,''
\newblock in {\em NIPS}, 2015, pp. 1225--1233.

\bibitem{bojarski17a}
M.~Bojarski, A.~Choromanska, K.~Choromanski, F.~Fagan, C.~Gouy-Pailler,
  A.~Morvan, N.~Sakr, T.~Sarlos, and J.~Atif,
\newblock ``{Structured adaptive and random spinners for fast machine learning
  computations},''
\newblock in {\em AISTATS}, 2017, pp. 1020--1029.

\bibitem{SH08}
Y.~Weiss, A.~Torralba, and R.~Fergus,
\newblock ``Spectral hashing.,''
\newblock in {\em NIPS}, 2008, pp. 1753--1760.

\bibitem{SHL12}
J.~Wang, S.~Kumar, and S.~Chang,
\newblock ``Semi-supervised hashing for large-scale search,''
\newblock {\em IEEE Trans. Pattern Anal. Mach. Intell.}, , no. 12, pp.
  2393--2406, 2012.

\bibitem{ITQ2013}
Y.~Gong, S.~Lazebnik, A.~Gordo, and F.~Perronnin,
\newblock ``Iterative quantization: A procrustean approach to learning binary
  codes for large-scale image retrieval,''
\newblock {\em IEEE Transactions on Pattern Analysis and Machine Intelligence},
  , no. 12, pp. 2916--2929, 2013.

\bibitem{IsoHash2012}
W.~Kong and W.~Li,
\newblock ``Isotropic hashing,''
\newblock in {\em NIPS}, pp. 1646--1654. 2012.

\bibitem{CBEICML14}
F.~Yu, S.~Kumar, Y.~Gong, and S.~Chang,
\newblock ``Circulant binary embedding,''
\newblock in {\em ICML}, 2014.

\bibitem{Jegou2010}
H.~J\'egou, M.~Douze, C.~Schmid, and P.~P\'erez,
\newblock ``Aggregating local descriptors into a compact image
  representation,''
\newblock in {\em CVPR}, 2010, pp. 3304--3311.

\bibitem{LengWC0L15}
C.~Leng, J.~Wu, J.and~Cheng, X.~Bai, and H.~Lu,
\newblock ``Online sketching hashing.,''
\newblock in {\em CVPR}, 2015, pp. 2503--2511.

\bibitem{surveyHashing16}
J.~Wang, W.~Liu, S.~Kumar, and S.~Chang,
\newblock ``Learning to hash for indexing big data - a survey,''
\newblock {\em Proceedings of the IEEE}, , no. 1, pp. 34--57, 2016.

\bibitem{SKLSH09}
M.~Raginsky and S.~Lazebnik,
\newblock ``Locality-sensitive binary codes from shift-invariant kernels,''
\newblock in {\em NIPS}, pp. 1509--1517. 2009.

\bibitem{KLSH11}
K.~Grauman and B.~Kulis,
\newblock ``Kernelized locality-sensitive hashing,''
\newblock {\em IEEE Transactions on Pattern Analysis and Machine Intelligence},
  pp. 1092--1104, 2011.

\bibitem{Liu11hashingwith}
W.~Liu, J.~Wang, and S.~Chang,
\newblock ``Hashing with graphs,''
\newblock in {\em ICML}, 2011.

\bibitem{Lee2012SphericalH}
Y.~Lee,
\newblock ``Spherical hashing,''
\newblock in {\em CVPR}, 2012, pp. 2957--2964.

\bibitem{DisGraphHashingNIPS2014}
W.~Liu, C.~Mu, S.~Kumar, and S.~Chang,
\newblock ``Discrete graph hashing,''
\newblock in {\em NIPS}, pp. 3419--3427. 2014.

\bibitem{Raziperchikolaei16}
R.~Raziperchikolaei and M.~{\'{A}}. Carreira{-}Perpi{\~{n}}{\'{a}}n,
\newblock ``Optimizing affinity-based binary hashing using auxiliary
  coordinates,''
\newblock in {\em NIPS}, 2016, pp. 640--648.

\bibitem{supervisedHK12}
W.~Liu, J.~Wang, R.~Ji, Y.~Jiang, and S.~Chang,
\newblock ``Supervised hashing with kernels,''
\newblock in {\em CVPR}, 2012, pp. 2074--2081.

\bibitem{LaiPLY15}
H.~Lai, Y.~Pan, Y.~Liu, and S.~Yan,
\newblock ``Simultaneous feature learning and hash coding with deep neural
  networks,''
\newblock in {\em CVPR}, 2015, pp. 3270--3278.

\bibitem{Chen2015CNN}
W.~Chen, J.~T. Wilson, S.~Tyree, K.~Q. Weinberger, and Y.~Chen,
\newblock ``Compressing neural networks with the hashing trick,''
\newblock in {\em ICML}, 2015, pp. 2285--2294.

\bibitem{HDDleng15}
C.~Leng, J.~Wu, J.~Cheng, X.~Zhang, and H.~Lu,
\newblock ``Hashing for distributed data,''
\newblock in {\em ICML}, 2015, pp. 1642--1650.

\bibitem{Huang2013OH}
L.~Huang, Q.~Yang, and W.~Zheng,
\newblock ``Online hashing,''
\newblock in {\em IJCAI}, 2013, pp. 1422--1428.

\bibitem{CakirHBS17}
F.~{\c{C}}akir, K.~He, S.A. Bargal, and S.~Sclaroff,
\newblock ``{MIH}ash: Online hashing with mutual information,''
\newblock in {\em ICCV}, Oct 2017.

\bibitem{OPAST2000}
K.~Abed-Meraim, A.~Chkeif, and Y.~Hua,
\newblock ``Fast orthonormal past algorithm,''
\newblock {\em IEEE Signal Processing Letters}, , no. 3, pp. 60 -- 62, 2000.

\bibitem{FengXY13}
J.~Feng, H.~Xu, and S.~Yan,
\newblock ``Online robust pca via stochastic optimization.,''
\newblock in {\em NIPS}, 2013, pp. 404--412.

\bibitem{YangX15a}
W.~Yang and H.~Xu,
\newblock ``Streaming sparse principal component analysis.,''
\newblock in {\em ICML}, 2015, pp. 494--503.

\bibitem{Golub2000}
G.~H. Golub and H.~A. van~der Vorst,
\newblock ``Eigenvalue computation in the 20th century,''
\newblock {\em Journal of Computational and Applied Mathematics}, , no. 1–2,
  pp. 35 -- 65, 2000,
\newblock Numerical Analysis 2000. Vol. III: Linear Algebra.

\bibitem{IR08}
C.D. Manning, P.~Raghavan, and H.~Sch{ü}tze,
\newblock {\em Introduction to Information Retrieval},
\newblock Cambridge University Press, 2008.

\end{thebibliography}

\end{document}